%% file: main.tex
\documentclass[runningheads]{llncs}
\usepackage[english]{babel}
\usepackage{graphicx}
\usepackage{amssymb}
\usepackage{amsmath}
\usepackage[table]{xcolor}
\usepackage{booktabs}
\usepackage{siunitx}
\usepackage{cite}
\usepackage{multirow}
\usepackage{tabularx}
\usepackage{rotating}
\usepackage{hyperref}
\usepackage{mymacros}
\usepackage{adjustbox}
\usepackage{color,soul}
\usepackage{eurosym}
\usepackage{paralist}

\usepackage{subcaption}

\addto\extrasenglish{%
}

\begin{document}
%
%
\title{Explainable Predictive Decision Mining for Operational Support}

\titlerunning{Explainable Predictive Decision Mining}
%
\author{Gyunam Park\inst{1}
\and
Aaron Küsters\inst{2}
\and
Mara Tews\inst{2}
\and
Cameron Pitsch\inst{2}
\and
Jonathan Schneider\inst{2}
\and
Wil M. P. van der Aalst\inst{1}
}
%
\authorrunning{Park et al.}
%
\institute{Process and Data Science Group (PADS), RWTH Aachen University, Germany \\ \email{\{gnpark,wvdaalst\}@pads.rwth-aachen.de}
\and
RWTH Aachen University, Germany \\ \email{\{aaron.kuesters,mara.tews,cameron.pitsch,lennart.schneider\}@rwth-aachen.de}
}
\maketitle              
\begin{abstract}
Several decision points exist in business processes (e.g., whether a purchase order needs a manager's approval or not), and different decisions are made for different process instances based on their characteristics (e.g., a purchase order higher than \euro500 needs a manager approval).
Decision mining in process mining aims to describe/predict the routing of a process instance at a decision point of the process.
By predicting the decision, one can take proactive actions to improve the process.
For instance, when a bottleneck is developing in one of the possible decisions, one can predict the decision and bypass the bottleneck.
However, despite its huge potential for such operational support, existing techniques for decision mining have focused largely on describing decisions but not on predicting them, deploying decision trees to produce logical expressions to explain the decision.
In this work, we aim to enhance the predictive capability of decision mining to enable proactive operational support by deploying more advanced machine learning algorithms.
Our proposed approach provides explanations of the predicted decisions using SHAP values to support the elicitation of proactive actions.
We have implemented a Web application to support the proposed approach and evaluated the approach using the implementation.
\end{abstract}

\input{Sections/1-Intro}
\input{Sections/2-Related}
\input{Sections/3-Background}
\input{Sections/4-Approach}
\input{Sections/5-Implementation}
\input{Sections/6-Evaluation}
\input{Sections/7-Conclusion}

\section*{Acknowledgment}
The authors would like to thank the Alexander von Humboldt (AvH) Stiftung for funding this research.

\bibliographystyle{splncs04}
\bibliography{mybib}

\end{document}

%% file: Sections/1-Intro.tex
\section{Introduction}
A process model represents the control-flow of business processes, explaining the routing of process instances.
It often contains decision points, e.g., XOR-split gateway in BPMN.
The routing in such decision points depends on the data attribute of the process instance.
For instance, in a loan application process, the assessment of a loan application depends on the amount of the loan, e.g., if the amount is higher than \euro$5000$, it requires \textit{advanced assessment} and, otherwise, \textit{simple assessment}.

Decision mining in process mining aims to discover a decision model that represents the routing in a decision point of a business process~\cite{DBLP:conf/sac/LeoniA13}.
The discovered decision model can be used for 1) describing how decisions have been made and 2) predicting how decisions will be made for future process instances.
While the focus has been on the former in the literature, the latter is essential to enable proactive actions to actually improve business processes~\cite{park_action-oriented_2022}.
Imagine we have a bottleneck in \textit{advanced assessment} due to, e.g., the lack of resources.
By predicting the decision of a future loan application, we can take proactive action (e.g., suggesting to lower the loan amount to conduct \textit{simple assessment}), thus facilitating the process.

\begin{figure}[!htb]
    \centering
    \includegraphics[width=1\linewidth]{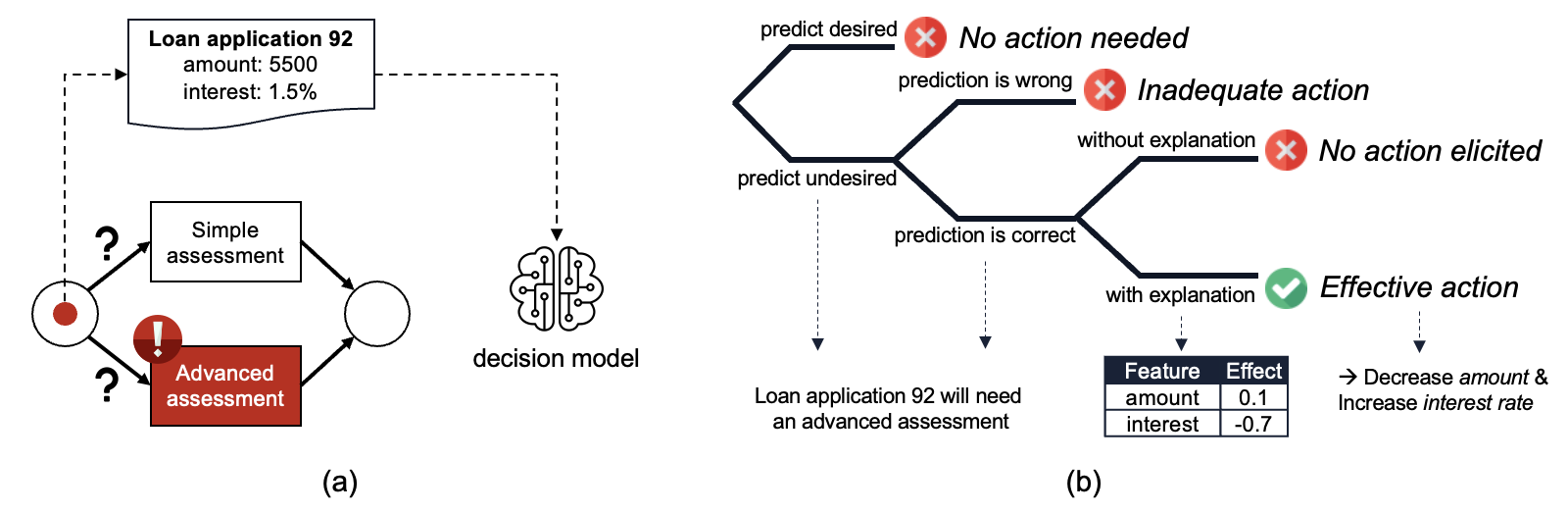}
    \caption{(a) Decision point in a process model. (b) Different Scenarios showing that decision mining needs to be predictive and descriptive to enable operational support.}
    \label{fig:intro}
\end{figure}

To enable such operational support, a decision model needs to be both 1) predictive (i.e., the model needs to provide reliable predictions on undesired/ decisions) and 2) descriptive (i.e., domain experts should be able to interpret how the decision is made to elicit a proactive action).
\autoref{fig:intro} demonstrates these requirements.
\autoref{fig:intro}(a) shows a decision point in a loan application process, and there is a bottleneck in \textit{advanced assessment}.
Our goal is to accurately predict that a loan application with the amount of \euro$5500$ and interest of $1.5$\% needs \textit{advanced assessment}, which is undesired due to the bottleneck, and recommend actions to avoid the bottleneck.
\autoref{fig:intro}(b) shows four different scenarios.
First, if we predict a desired decision (i.e., predicting the simple assessment), no action is required since the simple assessment has no operational issues.
Second, if we predict an undesired prediction incorrectly (e.g., incorrectly predicting the advanced assessment), we recommend an inadequate action. 
Third, if we predict the undesired decision correctly but no explanations are provided, no action can be elicited due to the lack of explanations.
Finally, if we predict the undesired decision, and the corresponding explanations are provided (e.g., the amount/interest of the loan has a positive/negative effect on the probability of conducting the advanced assessment, respectively), we can come up with relevant actions (e.g., lowering the amount or increasing the interest rate).

Existing work has focused on the descriptive capability of decision models by deploying highly interpretable machine learning algorithms such as decision trees~\cite{DBLP:conf/bpm/RozinatA06, DBLP:conf/caise/MannhardtLRA16, DBLP:conf/sac/LeoniA13}.
However, it leads to limited predictive capability due to the limitation of decision trees, such as overfitting and instability (i.e., adding a new data point results in regeneration of the overall tree)~\cite{DBLP:journals/tsmc/SafavianL91}.
In this work, we aim to enhance the predictive capabilities of decision mining, while providing explanations of predicted decisions.
To this end, we estimate the decision model by using machine learning algorithms such as support vector machines, random forests, and neural networks.
Next, we produce explanations of the prediction of the decision model by using SHAP values.

We have implemented the approach as a standalone web application.
Using the implementation, we have evaluated the accuracy of predicted decisions using real-life event logs.
Moreover, we have evaluated the reliability of explanations of predicted decisions by conducting controlled experiments using simulation models.

This paper is structured as follows. First, we discuss related work on decision mining and explainability in \autoref{sec:related}. Next, we introduce process models and event logs in \autoref{sec:preliminaries}. In \autoref{sec:approach}, we provide our proposed approach. In \autoref{sec:imple}, we explain the implementation of a web application based on the approach. \autoref{sec:evaluation} evaluates the approach based on the implementation using simulated and real-life event logs. We conclude this paper in \autoref{sec:conclusion}.

%% file: Sections/2-Related.tex
\section{Related Work}\label{sec:related}
Several approaches have been proposed to learn decision models from event logs.
Rozinat et al.~\cite{DBLP:conf/bpm/RozinatA06} suggest a technique based on Petri nets.
It discovers a Petri net from an event log, identifies decision points, and employs classification techniques to determine decision rules.
De Leoni et al.~\cite{DBLP:conf/sac/LeoniA13} extend \cite{DBLP:conf/bpm/RozinatA06} by dealing with invisible transitions of a Petri net and non-conforming process instances using \emph{alignments}.
These methods assume that decision-making is deterministic and all factors affecting decisions exist in event logs.
To handle non-determinism and incomplete information, Mannhardt et al.~\cite{DBLP:conf/caise/MannhardtLRA16} propose a technique to discover overlapping decision rules.
In~\cite{DBLP:conf/bpm/BazhenovaW15}, a framework is presented to derive decision models using Decision Model and Notation (DMN) and BPMN.
All existing approaches deploy decision trees due to their interpretability.
To the best of our knowledge, no advanced machine learning algorithms have been deployed to enhance the predictive capabilities of decision models along with explanations.

Although advanced machine learning approaches provide more accurate predictions compared to conventional white-box approaches, they lack explainability due to their black-box nature.
Recently, various approaches have been proposed to explain such black-box models.
Gilpin et al.~\cite{DBLP:journals/corr/abs-1806-00069} provide a systematic literature survey to provide an overview of explanation approaches.
The explanation approaches are categorized into \textit{global} and \textit{local} methods.
First, global explanation approaches aim to describe the average behavior of a machine learning model by analyzing the whole data.
Such approaches include Partial Dependence Plot (PDP)~\cite{DBLP:journals/corr/abs-1805-04755}, Accumulated Local Effects (ALE) Plot~\cite{https://doi.org/10.48550/arxiv.1612.08468}, and global surrogate models~\cite{DBLP:journals/corr/abs-1711-09784}.
Next, local explanation approaches aim to explain individual predictions by individually examining the instances.
Such approaches include Individual Conditional Expectation (ICE)~\cite{https://doi.org/10.48550/arxiv.1309.6392}, Local Surrogate (LIME)~\cite{DBLP:conf/kdd/Ribeiro0G16}, and Shapley Additive Explanations (SHAP)~\cite{DBLP:conf/nips/LundbergL17}.
In this work, we use SHAP to explain the predictions produced by decision models due to its solid theoretical foundation in game theory and the availability of global interpretations by combining local interpretations~\cite{DBLP:conf/nips/LundbergL17}.

%% file: Sections/3-Background.tex
\section{Preliminaries}\label{sec:preliminaries}
Given a set $X$, we denote the set of all multi-sets over $X$ with $\bag(X)$.
$f{\restriction}_{X}$ is the function projected on $X$: $dom(f{\restriction}_{X})=dom(f) \cap X$ and $f{\restriction}_{X}(x)=f(x)$ for $x \in dom(f\restriction_{X})$. 

\subsection{Process Models}
Decision mining techniques are independent of the formalism representing process models, e.g., BPMN, YWAL, and UML-activity diagrams.
In this work, we use Petri nets as the formalism to model the process.

First, a Petri net is a directed bipartite graph of places and transitions. 
A labeled Petri net is a Petri net with the transitions labeled.

\begin{definition}[Labeled Petri Net]\label{def:lpn}
Let $\univ{act}$ be the universe of activity names.
A labeled Petri net is a tuple $N {=} (P,T,F,l)$ with $P$ the set of places, $T$ the set of transitions,
$P \cap T {=} \emptyset$, $F\subseteq (P \times T) \cup (T \times P)$ the flow relation, and $l \in T \not\rightarrow \univ{act}$ a labeling function.
\end{definition}

\begin{figure}[!htb]
    \centering
    \includegraphics[width=1\linewidth]{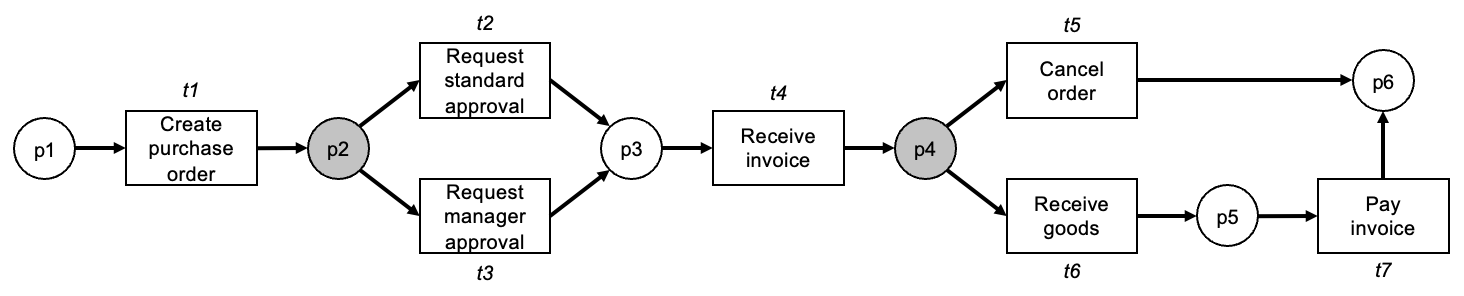}
    \caption{An example of Petri nets highlighted with decision points}
    \label{fig:petri-net}
\end{figure}

\autoref{fig:petri-net} shows a Petri net, $N_1=(P_1,T_1,F_1,l_1)$, where $P_1 = \{ p1,\dots, p6 \}$, $T_1 = \{ t1,\dots,t7\}$, $F_1 = \{(p1,t1),(t1,p2),\dots \}$, $l_1(t1)
= \textit{Create purchase order}$, $l_1(t2) = \textit{Request standard approval}$, etc.

The state of a Petri net is defined by its marking.
A marking $M_{N} \in \bag(P)$ is a multiset of places.
For instance, $M_{N_1}=[p1]$ represents a marking with a token in $p1$.
A transition $tr \in T$ is \textit{enabled} in marking $M_{N}$ if its input places contain at least one token.
The enabled transition may \textit{fire} by removing one token from each of the input places and producing one token in each of the output places.
For instance, $t1$ is \textit{enabled} in $M_{N_1}$ and \textit{fired} by leading to $M'_{N_1}=[p2]$.

\begin{definition}[Decision Points]\label{def:lpn}
Let $N {=} (P,T,F,l)$ be a labeled Petri net.
For $p \in P$, $\post p=\{t \in T \mid (p,t) \in F\}$ denotes it outgoing transitions.
$p \in P$ is a decision point $if$ $ | \post p | >1$.
\end{definition}

For instance, $p2$ is a decision point in $N_1$ since $\post{p2} = \{t2,t3\}$ and $| \post{p2} | > 1$.

\subsection{Event Logs}

\begin{definition}[Event Logs]
Let $\univ{event}$ be the universe of events, $\univ{attr}$ the universe of attribute names ($\{case,act,time,res\} \subseteq \univ{attr}$), and $\univ{val}$ the universe of attribute values.
An event log is a tuple $L=(E,\pi)$ with $E \subseteq \univ{event}$ as the set of events and $\pi \in E \rightarrow (\univ{attr} \nrightarrow \univ{val})$ as the value assignments of the events.
\end{definition}

\autoref{tab:eventlog} shows a part of an event log $L_1{=}(E_1,\pi_1)$. $e_1 \in E_1$ represents the event in the first row, i.e., $\pi_1(e_1)(case)=\textit{PO92}$, $\pi_1(e_1)(act)=\textit{Create Purchase Order}$, $\pi_1(e_1)(time)=\text{\scriptsize 09:00 05.Oct.2022}$, $\pi_1(e_1)(res)=\textit{Adams}$, $\pi_1(e_1)(\textit{total-price})=1000$, and $\pi_1(e_1)(\textit{vendor})=\textit{Apple}$.

\begin{table}[t!]
\centering
\caption{An Example of event logs}
\label{tab:eventlog}
\resizebox{0.9\textwidth}{!}{%
\begin{tabular}{|c|c|c|c|c|c|}
\hline
case id & activity               & timestamp         & resource & total-price & vendor  \\ \hline
PO92    & Create Purchase Order  & 09:00 05.Oct.2022 & Adams    & 1000        & Apple   \\ \hline
PO92    & Request Standard Order & 11:00 07.Oct.2022 & Pedro    & 1000        & Apple   \\ \hline
PO93    & Create Purchase Order  & 13:00 07.Oct.2022 & Peter    & 1500        & Samsung \\ \hline
\dots   & \dots                  & \dots             & \dots    & \dots       & \dots   \\ \hline
\end{tabular}%
}
\end{table}

%% file: Sections/4-Approach.tex
\section{Explainable Predictive Decision Mining}\label{sec:approach}
\begin{figure}[!htb]
    \centering
    \includegraphics[width=1\linewidth]{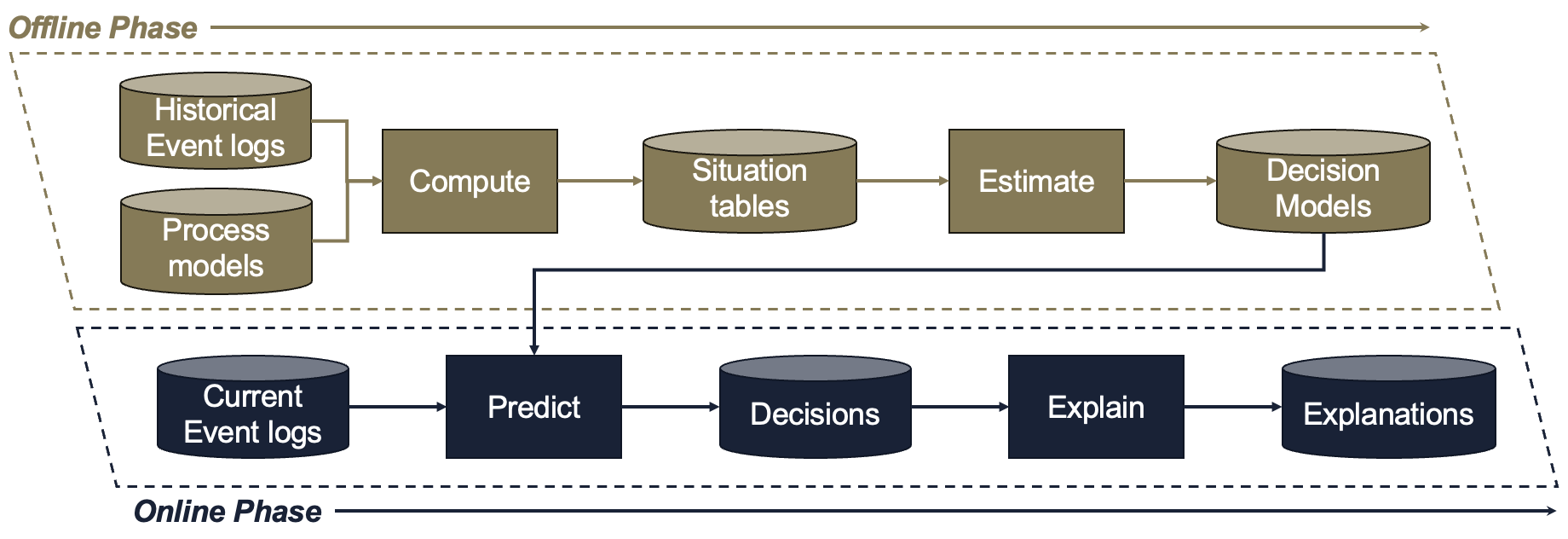}
    \caption{An overview of the proposed approach}
    \label{fig:approach}
\end{figure}

In this section, we introduce an approach to explainable predictive decision mining.
As shown in \autoref{fig:approach}, the proposed approach consists of two phases: \textit{offline} and \textit{online} phases.
The former aims to derive decision models of decision points, while the latter aims at predicting decisions for running process instances along with explanations.
In the offline phase, we compute \emph{situation tables} based on historical event logs and estimate decision models using the \emph{situation tables}.
In the online phase, we predict decisions for ongoing process instances and explain the decision.

\subsection{Offline Phase}
First, we compute situation tables from event logs.
Each record in a situation table consists of features (e.g., total price of an order) and a decision in a decision point (e.g., $t2$ at decision point $p2$ in \autoref{fig:petri-net}), describing how the decision has been historically made (e.g., at decision point $p2$ in \autoref{fig:petri-net}, standard approval (i.e., $t2$) was performed when the total price of an order is \euro$1000$).

\begin{definition}[Situation Table]
Let $\univ{feature}$ be the universe of feature names and $\univ{fmap} = \univ{feature} \nrightarrow \univ{val}$ the universe of feature mappings.
Let $N {=} (P,T,F,l)$ be a labeled Petri net and $p \in P$ a decision point.
$sit_{p} \in \univ{L} \rightarrow \bag(\univ{fmap} \times \post p)$ maps event logs to situation tables (i.e., multi-sets of feature mappings and decisions).
$S_{\mi{p}}{=}\{sit_{\mi{p}}(L) \mid L \in \univ{L} \}$ denotes the set of all possible situation tables of $p$.
\end{definition}

\begin{figure}[!htb]
    \centering
    \includegraphics[width=1\linewidth]{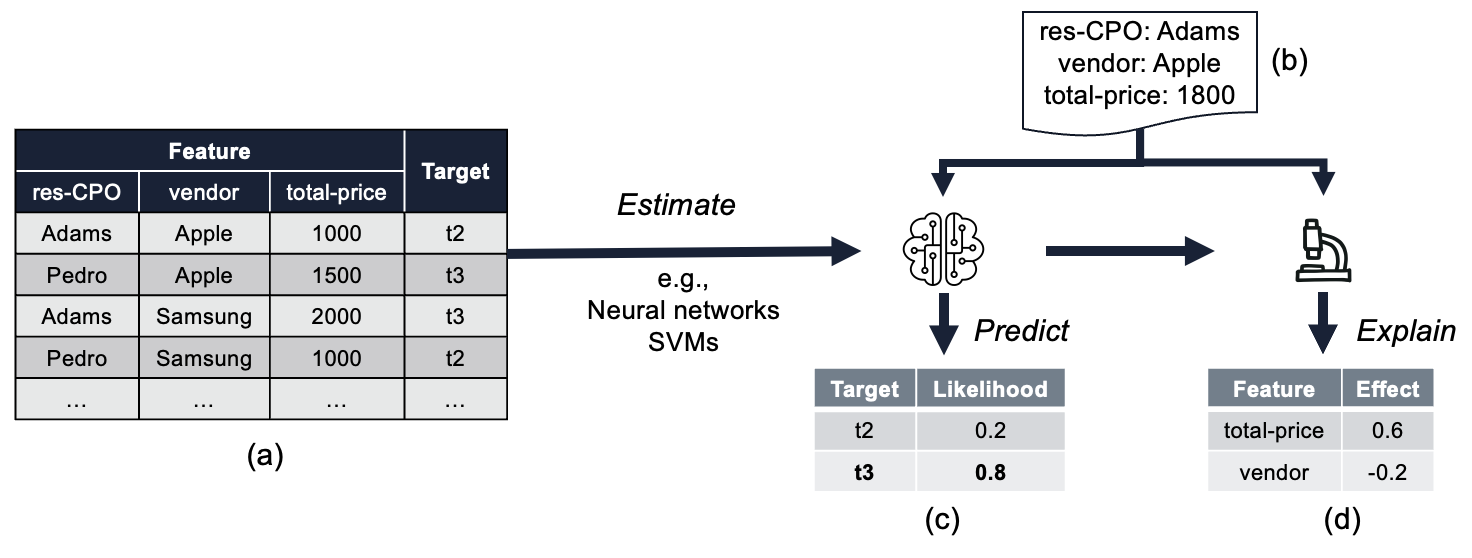}
    \caption{An example of the proposed approach}
    \label{fig:approach-example}
\end{figure}

The table in \autoref{fig:approach-example}(a) represents a situation table of $p2$ in \autoref{fig:petri-net} derived from the event log depicted in \autoref{tab:eventlog}.
For instance, the first row in \autoref{fig:approach-example}(a) describes that \textit{request standard approval} ($t2$) was executed when human resource \textit{Adams} performed \textit{create purchase order} (i.e., \textit{res-CPO}) for the order of \euro$1000$ (i.e., \textit{total-price}) with \textit{Apple} (i.e., \textit{vendor}).
Formally, $s_1=(fmap_1,t2) \in sit_{\mi{p2}}(L1)$ where $fmap_1 \in \univ{fmap}$ such that  $fmap_1=\{(\textit{res-CPO},\textit{Adams}),(\textit{vendor},\textit{Apple}),(\textit{total-} \allowbreak \textit{price},1000)\}$.
Note that, $s_1$ corresponds to event $e_2$ in \autoref{tab:eventlog} and $fmap_1$ is derived from all historical events of $PO92$.

A decision model provides the likelihood of each transition in a decision point based on a given feature, e.g., when the total price of an order (i.e., feature) is \euro$1800$, standard approval will be performed with the likelihood of $0.2$ and manager approval with the likelihood of $0.8$.

\begin{definition}[Decision Model]
Let $N {=} (P,T,F,l)$ be a labeled Petri net and $p \in P$ a decision point.
Let $dmap_{\mi{p}} \in \post p \rightarrow [0,1]$ be a decision mapping that maps decisions to likelihoods such that the sum of all likelihoods adds up to 1, i.e., $\Sigma_{p' \in \post p} dmap_{\mi{p}}(p') = 1$.
$D_{\mi{p}}$ denotes the set of all possible decision mappings.
$DM_{\mi{p}} \in \univ{fmap} \rightarrow D_{\mi{p}}$ is the set of all possible decision models of $p$ that map feature mappings to decision mappings.
\end{definition}

We estimate decision models based on situation tables.

\begin{definition}[Estimating Decision Models]
Let $N {=} (P,T,F,l)$ be a labeled Petri net and $p \in P$ a decision point.
$estimate_{\mi{p}} \in S_{\mi{p}} \rightarrow DM_{\mi{p}}$ is a function estimating a decision model from a situation table.
\end{definition}

The estimation function can be built using many machine learning algorithms such as neural networks, support vector machines, random forests, etc.

\subsection{Online Phase}

Using the decision model derived from the offline phase, we predict the decision of a running process instance and explain the prediction.
Using the feature of a running process instance depicted in \autoref{fig:approach-example}(b), a decision model may produce the prediction shown in \autoref{fig:approach-example}(c), leading to the final decision of \textit{request manager approval} that has the highest likelihood.
Next, we compute an explanation for the decision (i.e., the effect of each feature on the prediction) as shown in \autoref{fig:approach-example}(d), e.g., \textit{total-price} has a positive effect of $0.6$ while \textit{vendor} has a negative effect of $0.2$.
In other words, \textit{total-price} increases the likelihood of predicting the decision of \textit{request manager approval} by the magnitude of $0.6$ and \textit{vendor} decreases it by the magnitude of $0.2$, respectively.

In this work, we use SHAP values~\cite{DBLP:conf/nips/LundbergL17} to provide explanations of decisions.
SHAP values are based on \emph{Shapley} values.
The concept of \emph{Shapley} values comes from \emph{game theory}.
It defines two elements: a game and some players.
In the context of predictions, the game is to reproduce the outcome of the model, and the players are the features used to learn the model.
Intuitively, Shapley values quantify the amount that each player contributes to the game, and SHAP values quantify the contribution that each feature brings to the prediction made by the model.

\begin{definition}[Explaining Decisions]
Let $fmap \in \univ{fmap}$ be a feature mapping and $F=\{f_1,f_2,\dots,f_i,\dots\}=dom(fmap)$ denote the domain of $fmap$.
Let $N {=} (P,T,F,l)$ be a labeled Petri net, $p \in P$ a decision point, and $dm_{\mi{p}}$ a decision model.
Let $t \in \post p$ be a target transition.
The SHAP value of feature $f_i$ for predicting $t$ is defined as:
\begin{equation*}
    \psi_{f_i}^{\mi{t}}=\sum_{F' \subseteq F \setminus \{f_i\} } \frac{\card{F'}!(\card{F}-\card{F'}-1)!}{\card{F}!} (dm_{\mi{p}}(fmap{\restriction}_{F'\cup \{f_i\}})(t)-dm_{\mi{p}}(fmap{\restriction}_{F'})(t))
\end{equation*}
For $fmap$, $exp_{\mi{dm}_{p},\mi{t}}(fmap)=\{(f_1,\psi_{f_1}^{\mi{t}}),(f_2,\psi_{f_2}^{\mi{t}}),\dots\}$ is the explanation of predicting $t$ using $dm_{p}$.
\end{definition}

As shown in \autoref{fig:approach-example}(d), for feature mapping $fmap'$ described in \autoref{fig:approach-example}(b), the explanation of predicting $t3$ (i.e., request manager approval) using decision model $dm_{p2}'$ is $exp_{\mi{dm}_{p2}',\mi{t3}}(fmap')=\{(\textit{total-price},0.6),(\textit{vendor},-0.2)\}$.
In other words, \textit{total-price} has a positive effect with the magnitude of $0.6$ on the decision of $t3$ and \textit{vendor} has a negative effect with the magnitude of $0.2$.

Moreover, we can provide a global explanation of a decision model by aggregating SHAP values of multiple running instances.
For instance, by aggregating all SHAP values of \textit{total-price} for predicting $t3$, e.g., with the mean absolute value, we can compute the global effect of \textit{total price} to the prediction.

%% file: Sections/5-Implementation.tex
\section{Implementation}\label{sec:imple}
We have implemented a Web application to support the explainable decision mining with a dedicated user interface.
Source code and user manuals are available at \url{https://github.com/aarkue/eXdpn}.
The application comprises three functional components as follows.

\paragraph{\textbf{Discovering Process Models.}}
This component supports the discovery of process models based on inductive miner~\cite{DBLP:conf/apn/LeemansFA13}. 
The input is event data of the standard XES.
The discovered accepting Petri net is visualized along with its decision points.

\paragraph{\textbf{Decision Mining.}}
This component supports the computation of situation tables from event logs and the estimation of decision models from the computed situation table.
First, it computes situation tables with the following three types of features:
\begin{itemize}
    \item \textit{Case features}: Case features are on a case-level and used for predicting all decisions related to that case.
    \item \textit{Event features}: Event features are specific to an event and used for predicting decisions after the occurrence of the event.
    \item \textit{Performance features}: Performance features are derived from the log. It includes \textit{elapsed time of a case} (i.e., time duration since the case started) and \textit{time since last event} (i.e., time duration since the previous event occurred).
\end{itemize}

Next, the estimation of decision models uses the following machine learning algorithms: \emph{Random Forests}, \emph{XGBoost}, \emph{Support Vector Machines (SVMs)}, and \emph{Neural Networks}.

\begin{figure}[!htb]
    \centering
    \includegraphics[width=1\linewidth]{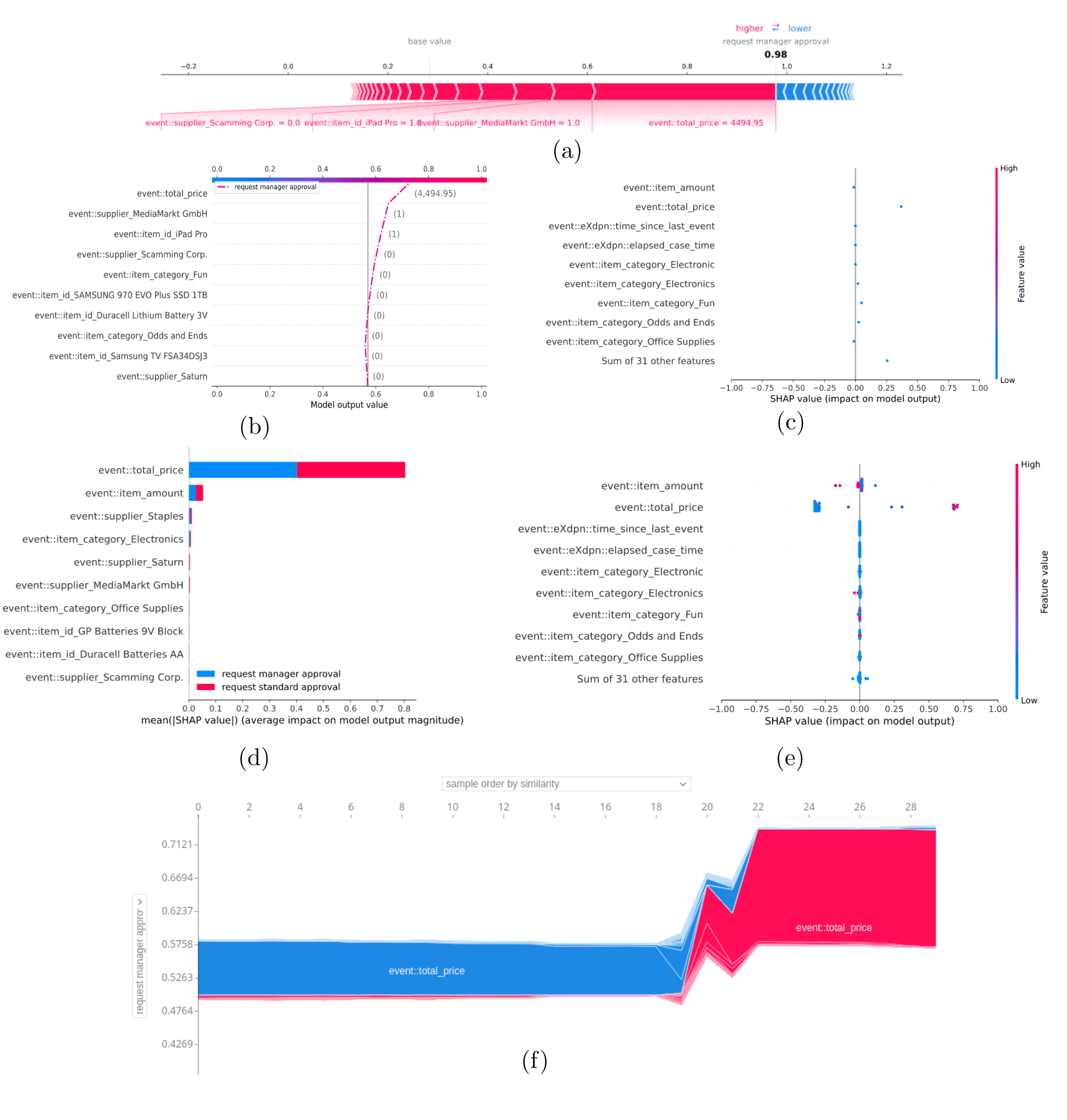}
    \caption{
    \textbf{Local explanations}: (a) Force plot, (b) Decision plot, and (c) Beeswarm plot explain how the model arrived at the decision of a running instance (i.e., \textit{request manager approval} with the likelihood of $0.98$).
    For instance, (a) visualizes the positive (red-colored) and negative (blue-colored) features with increasing magnitudes.
    \textbf{Global explanations}: (d) Bar plot, (e) Beeswarm plot, and (f) Force plot explain how the model arrived at the decision of all running instances (both on \textit{request standard approval} and \textit{request manager approval}).
    For instance, (d) visualizes the mean absolute SHAP value for each feature on predicting \textit{request standard approval} (blue-colored bar) and \textit{request manager approval} (red-colored bar), showing that total-price has the highest impact on both predictions.
    }
    \label{fig:viz}
\end{figure}

\paragraph{\textbf{Visualizing Decisions and Explanations}}
This component visualizes the F1 score of different machine learning algorithms and suggests the best technique based on the score.
Moreover, it visualizes the explanation of the decision both at local and global levels.
Local explanations are visualized with \textit{force plot} (cf. \autoref{fig:viz}(a)), \textit{decision plot} (cf. \autoref{fig:viz}(b)), and \textit{beeswarm plot} (cf. \autoref{fig:viz}(c)), whereas global explanations are visualized with \textit{bar plot} (cf. \autoref{fig:viz}(d)), \textit{force plot} (cf. \autoref{fig:viz}(e)), and \textit{beeswarm plot} (cf. \autoref{fig:viz}(f)).
We refer readers to \cite{shap} for the details of different plots.

%% file: Sections/6-Evaluation.tex
\section{Evaluation}\label{sec:evaluation}
In this section, we evaluate the approach by conducting experiments using the implementation.
Specifically, we are interested in answering the following research questions.
\begin{compactitem}
    \item RQ1: Does the advanced machine learning algorithm efficiently predict the decisions?
    \item RQ2: Does the approach provides reliable explanations for the predictions?
\end{compactitem}

\subsection{RQ1: Prediction Accuracy}
In order to answer RQ1, we conduct experiments using real-life event logs: Business Process Intelligence Challenge (BPIC) 2012\footnote{\url{doi:10.4121/uuid:3926db30-f712-4394-aebc-75976070e91f}} and BPIC 2019\footnote{\url{doi.org/10.4121/uuid:d06aff4b-79f0-45e6-8ec8-e19730c248f1}}.
For each event log, we first discover a process model and determine decision points. Then we estimate different decision models for each decision point and compare the performance of the decision models using 5-fold cross-validation. To measure the performance of the decision model, we use F1 scores. Each model is instantiated with suitable, event-log-specific parameters, which have largely been obtained from a parameter grid search on each decision point as well as manual test runs.
For decision tree algorithms, we apply pruning steps to avoid too many splits that result in decision trees harder to interpret in practice due to their complexity.

\begin{table}[t!]
\centering
\caption{F1 scores of applying different machine learning algorithms in different decision points. The bold font shows the top two results in each decision point.}
\label{tab:quantitative}
\resizebox{1\textwidth}{!}{%
\begin{tabular}{|cc|cccccc|cccc|}
\hline
\multicolumn{2}{|c|}{Event Logs} &
  \multicolumn{6}{c|}{BPI Challenge 2012 (only Offers)} &
  \multicolumn{4}{c|}{BPI Challenge 2019 (filtered)} \\ \hline
\multicolumn{2}{|c|}{Decision point} &
  \multicolumn{1}{c|}{p4} &
  \multicolumn{1}{c|}{p6} &
  \multicolumn{1}{c|}{p12} &
  \multicolumn{1}{c|}{p14} &
  \multicolumn{1}{c|}{p16} &
  p19 &
  \multicolumn{1}{c|}{p3} &
  \multicolumn{1}{c|}{p4} &
  \multicolumn{1}{c|}{p8} &
  p11 \\ \hline
\multicolumn{1}{|c|}{\multirow{5}{*}{Algorithms}} &
  Decision Tree &
  \multicolumn{1}{c|}{0.6888} &
  \multicolumn{1}{c|}{0.7545} &
  \multicolumn{1}{c|}{0.7955} &
  \multicolumn{1}{c|}{0.9633} &
  \multicolumn{1}{c|}{\textbf{0.9612}} &
  \multicolumn{1}{c|}{0.9263} &
  \multicolumn{1}{c|}{0.9555} &
  \multicolumn{1}{c|}{0.9948} &
  \multicolumn{1}{c|}{0.8135} &
  \textbf{1.0000} \\ \cline{2-12} 
\multicolumn{1}{|c|}{} &
  XGBoost &
  \multicolumn{1}{c|}{\textbf{0.7189}} &
  \multicolumn{1}{c|}{\textbf{0.7897}} &
  \multicolumn{1}{c|}{\textbf{0.8004}} &
  \multicolumn{1}{c|}{0.9697} &
  \multicolumn{1}{c|}{\textbf{0.9612}} &
  \multicolumn{1}{c|}{\textbf{0.9407}} &
  \multicolumn{1}{c|}{\textbf{0.9632}} &
  \multicolumn{1}{c|}{0.9948} &
  \multicolumn{1}{c|}{\textbf{0.8293}} &
  \textbf{1.0000} \\ \cline{2-12}
\multicolumn{1}{|c|}{} &
  Support Vector Machine &
  \multicolumn{1}{c|}{0.7151} &
  \multicolumn{1}{c|}{0.7799} &
  \multicolumn{1}{c|}{\textbf{0.8023}} &
  \multicolumn{1}{c|}{\textbf{0.9701}} &
  \multicolumn{1}{c|}{\textbf{0.9612}} &
  \multicolumn{1}{c|}{\textbf{0.9414}} &
  \multicolumn{1}{c|}{\textbf{0.9649}} &
  \multicolumn{1}{c|}{\textbf{0.9950}} &
  \multicolumn{1}{c|}{0.8096} &
  0.9997 \\ \cline{2-12}
\multicolumn{1}{|c|}{} &
  Neural Network &
  \multicolumn{1}{c|}{\textbf{0.725}} &
  \multicolumn{1}{c|}{\textbf{0.8048}} &
  \multicolumn{1}{c|}{0.7955} &
  \multicolumn{1}{c|}{\textbf{0.9698}} &
  \multicolumn{1}{c|}{0.9607} &
  \multicolumn{1}{c|}{0.9317} &
  \multicolumn{1}{c|}{0.9583} &
  \multicolumn{1}{c|}{\textbf{0.9981}} &
  \multicolumn{1}{c|}{\textbf{0.8191}} &
  0.9949 \\ \hline
\end{tabular}%
}
\end{table}

\autoref{tab:quantitative} shows the F1 score of different machine learning algorithms in different real-life event logs\footnote{The experimental results are reproducible in \url{https://github.com/aarkue/eXdpn/tree/main/quantitative_analysis} along with the corresponding process model.}.
The top two scores for each decision point are highlighted with bold fonts.
\emph{XGBoost} shows good scores for all decision points except $p14$ in BPIC 2012 and $p4$ in BPIC 2019. 
The scores for \emph{Support Vector Machine} belong to the top two scores for most of the decisions except $p4$ and $p6$ in BPIC 2012 and $p8$ and $p11$ in BPIC 2019, whereas the ones for \emph{Neural Network} belong to the top two scores in $p4$, $p6$ and $p14$ in BPIC 2012 and $p4$ and $p8$ in BPIC 2019.
\emph{Decision Tree} shows the top two scores only for $p16$ in BPIC 2012 and $p11$ in BPIC 2019.

\subsection{RQ2: Reliability of Explanations}
\begin{figure}[!htb]
    \centering
    \includegraphics[width=1\linewidth]{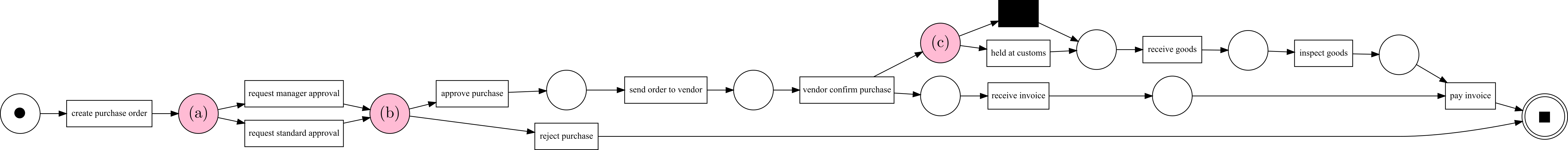}
    \caption{Petri net discovered from the simulated P2P event logs}
    \label{fig:exp:pn}
\end{figure}

To answer RQ2, we design a simulation model to simulate a Purchase-To-Pay (P2P) process using CPN tools~\cite{DBLP:conf/caise/ParkA21}.
The simulation model allows us to fully define the decision logic of decision points.
Based on the decision logic, we qualitatively evaluate if the generated explanation is reliable.
\autoref{fig:exp:pn} shows the Petri net discovered using inductive miner~\cite{DBLP:conf/apn/LeemansFA13} from an event log generated by the simulation model, with highlighted decision points.
Decision point (c) describes the decision of whether the purchase order is held at customs or not. 
The decision logic in the simulation model is as follows: \textit{If 1) a purchase order originates from outside the EU and 2) the base price per item is higher than \euro50, the order is held at customs.}

\begin{figure}[!htb]
     \centering
     \begin{subfigure}[b]{0.8\textwidth}
         \centering
         \includegraphics[width=1\textwidth]{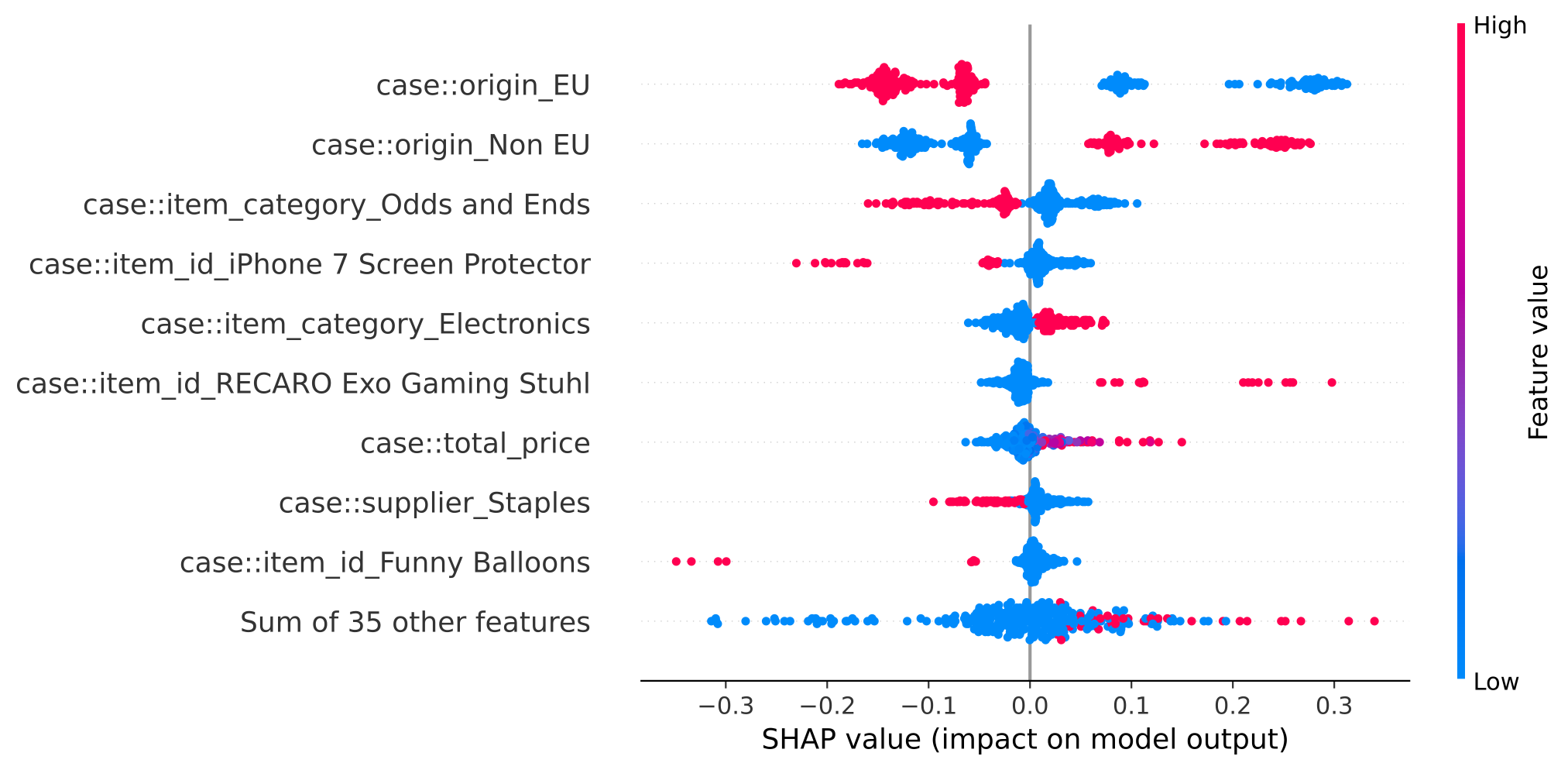}
         \caption{Beeswarm plot visualizing the impact of high or low feature values on the model probability of being held at customs.
         The Non-EU origin (high value of \texttt{origin\_Non EU}) has a strong positive impact on the probability of being held at customs.
         }
         \label{fig:exp:c:non_filtered}
     \end{subfigure}
     \begin{subfigure}[b]{0.8\textwidth}
         \centering
         \includegraphics[width=1\textwidth]{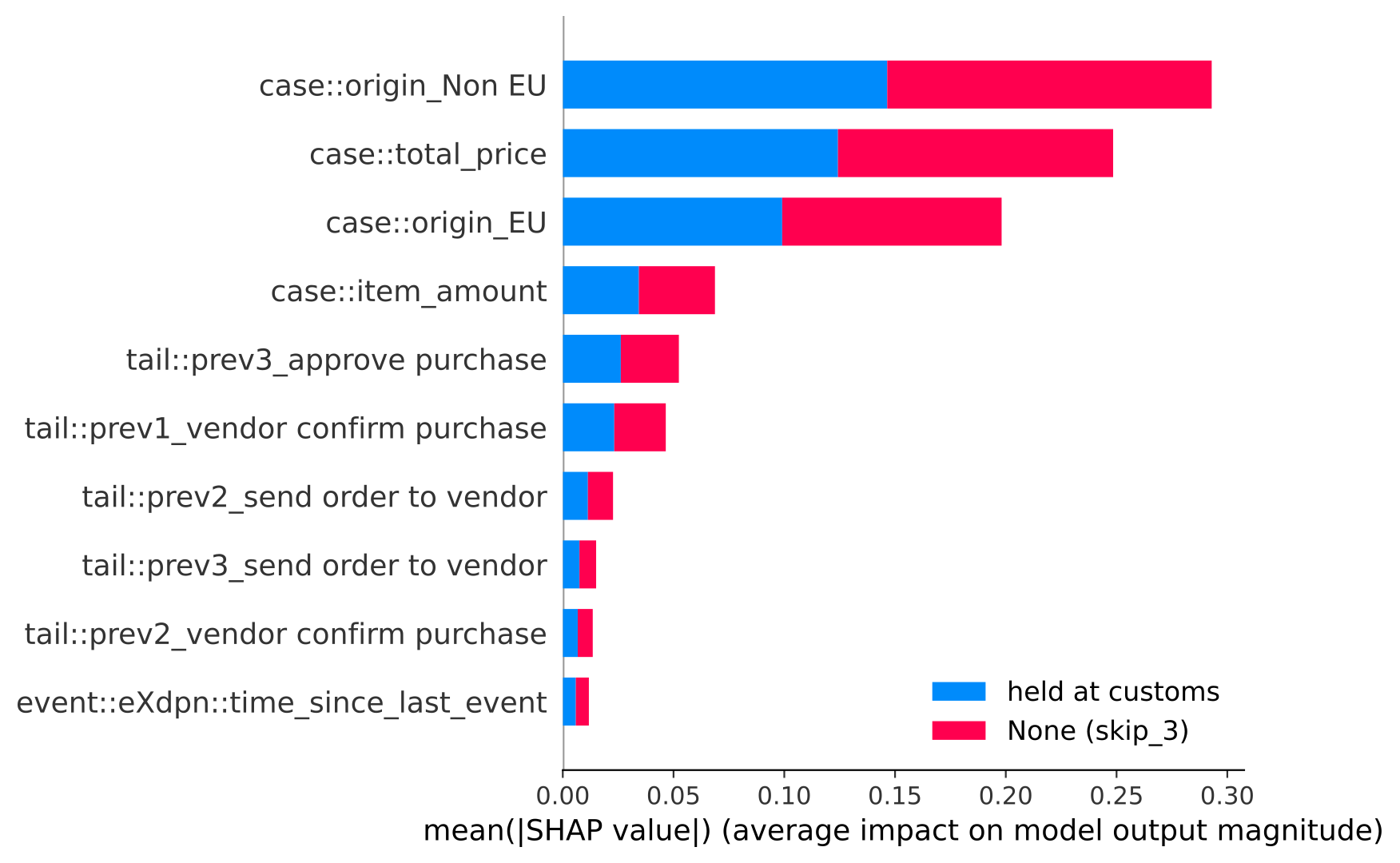}
         \caption{Bar plot visualizing the mean absolute SHAP value of each selected feature, per output class}
         \label{fig:exp:filtered_barplot}
     \end{subfigure}
     \caption{Qualitative Analysis showing the explanation plots of decision point (c) using a neural network model.}
     \label{fig:exp}
\end{figure}

The beeswarm plot in \autoref{fig:exp}(a) explains the decision at decision point (c).
The Non-EU origin (high value of \texttt{origin\_Non EU}) has a strong positive impact on the probability of being held at customs according to the decision model.
Moreover, the existence of items in category \texttt{Odds and Ends}, which have low base prices, has a negative impact on the probability, whereas the existence of items in category \texttt{Electronics}, which have high base prices, has a positive impact on the probability.
When the individual product names, categories, and vendors are excluded (see \autoref{fig:exp:filtered_barplot}), the four most impactful features that remain are exactly the ones used in the logic of the underlying simulation model: The EU or Non-EU origin, the total price and the number of items in the order.
Overall the decision logic as interpretable through the plots corresponds to the underlying logic applied in the simulation model, showing that the explanation obtained is reliable.

%% file: Sections/7-Conclusion.tex
\section{Conclusions}\label{sec:conclusion}
In this paper, we proposed an approach to explainable predictive decision mining.
In the offline phase of the approach, we derive decision models for different decision points.
In the online phase, we predict decisions for running process instances with explanations.
We have implemented the approach as a web application and evaluated the prediction accuracy using real-life event logs and the reliability of explanations with a simulated business process.

This paper has several limitations.
First, the explanation generated by the proposed approach is less expressive than the logical expression generated by traditional decision mining techniques.
Also, we abstract from the definition of features that can be used to construct the situation tables, focusing on explaining several possible features in the implementation.
In future work, we plan to extend the approach with a taxonomy of features to be used for the comprehensive construction of situation tables.
Moreover, we plan to connect the explainable predictive insights to actual actions to improve the process.